# A Novel comprehensive method for real time Video Motion Detection Surveillance

Sumita Mishra, Prabhat Mishra, Naresh K Chaudhary, Pallavi Asthana

**Abstract**— This article describes a comprehensive system for surveillance and monitoring applications. The development of an efficient real time video motion detection system is motivated by their potential for deployment in the areas where security is the main concern. The paper presents a platform for real time video motion detection and subsequent generation of an alarm condition as set by the parameters of the control system. The prototype consists of a mobile platform mounted with RF camera which provides continuous feedback of the environment. The received visual information is then analyzed by user for appropriate control action, thus enabling the user to operate the system from a remote location. The system is also equipped with the ability to process the image of an object and generate control signals which are automatically transmitted to the mobile platform to track the object.

**Index Terms**— Graphic User Interface, object tracking, Monitoring, Spying, Surveillance, video motion detection

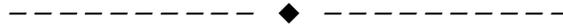

## 1 INTRODUCTION

Video Motion Detection Security Systems (VMDss) have been available for many years. Motion detection is a feature that allows the camera to detect any movement in front of it and transmit the image of the detected motion to the user. VMDss are based on the ability to respond to the temporal and/or spatial variations in contrast caused by movement in a video image. Several techniques for motion detection have been proposed, among them the three widely used approaches are background subtraction optical flow and temporal differencing. Background subtraction is the most commonly used approach in present systems. The principle of this method is to use a model of the background and compare the current image with a reference. In this way the foreground objects present in the scene are detected. Optical flow is an approximation of the local image motion and specifies how much each image pixel moves between adjacent images. It can achieve success of motion detection in the presence of camera motion or background changing. According to the smoothness constraint, the corresponding points in the two successive frames

————————————————
*Sumita Mishra is currently pursing doctoral degree in Electronics at DRML Avadh University, India and working as a lecturer in electronics and communication engineering department at Amity University, India*
*E mail: mishra.sumita@gmail.com*
*Prabhat Mishra is currently pursuing masters degree program in electronics and communication engineering in Amity University, India*

should not move more than a few pixels. For an uncertain environment, this means that the camera motion or background changing should be relatively small. Temporal differencing based on frame difference, attempts to detect moving regions by making use of the difference of consecutive frames (two or three) in a video sequence.

This method is highly adaptive to dynamic environments hence it is suitable for present application with certain modification. Presently advanced surveillance systems are available in the market at a very high cost. This paper aims at the low cost efficient security system having user friendly functional features which can also be controlled from a remote location. In addition the system can also be used to track the object of a predefined color rendering it useful for spying purposes.

## 2 HARDWARE SETUP

The proposed system comprises of two sections. The transmitter section consists of a computer , RS232 Interface, microcontroller, RF Transmitter, RF video receiver. The Receiver section consists of a Mobile Platform, RF receiver, microcontroller, RF camera, motor driver, IR LEDs. The computer at the transmitter section which receives the visual information from camera mounted on mobile platform works as control centre. Another function of control centre is to act as the web server that enables access to system from a remote location by using internet. The control centre is also responsible for transmitting the necessary control signal to the mobile platform.

## 3 MODES OF OPERATION

The system can operate in four independent modes.

## 3.1 PC Controlled Mode

In this mode the mobile platform is directly controlled by control centre using a visual GUI program developed using Microsoft Visual Studio 6.0(Visual Basic programming language). The user can control the mobile platform after analyzing the received video.

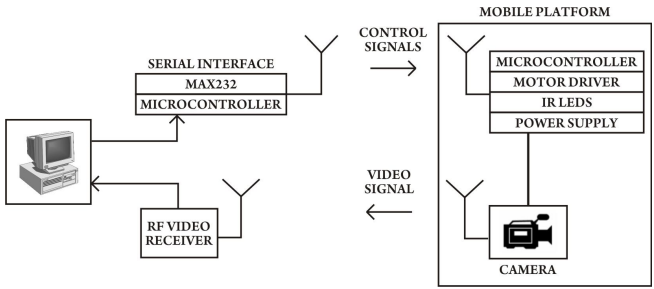

**Fig 1. Setup for PC Controlled Mode**

## 3.2 Internet Controlled Mode

This mode is an extension to the PC Controlled mode where client-server architecture is incorporated. This mode enables an authorized client computer to control the mobile platform from a remote location via internet. Client logs onto the control centre which provides all control tools for maneuvering the mobile platform. Instant images of the environment transmitted from the camera mounted on the mobile platform are used to generate appropriate control signals.

**Fig 2. Setup for Internet Controlled mode**

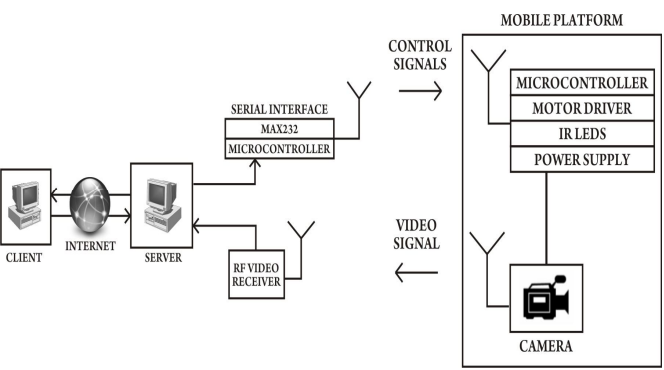

## 3.3 Tracing Mode

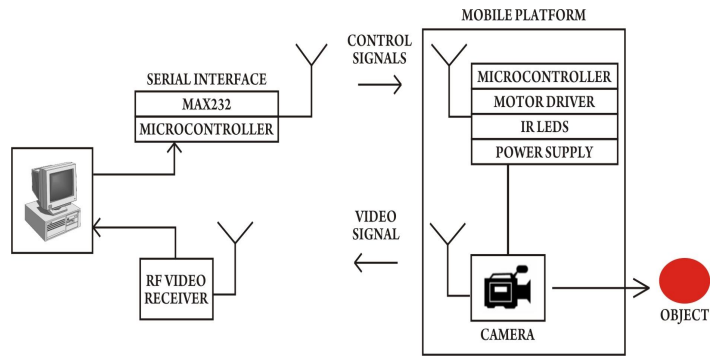

**Fig 3. Setup for tracing mode**

In this mode the system is made to follow the object whose color information has been stored at the control centre in program developed in MATLAB. Basically the program performs the image processing of the object and generates the control signals in order to make the mobile platform to trace the object.

## 3.4 Motion Detection Mode

In this mode the platform is made to focus on a particular object whose security is our concern. The mobile platform transmits the visual information of the

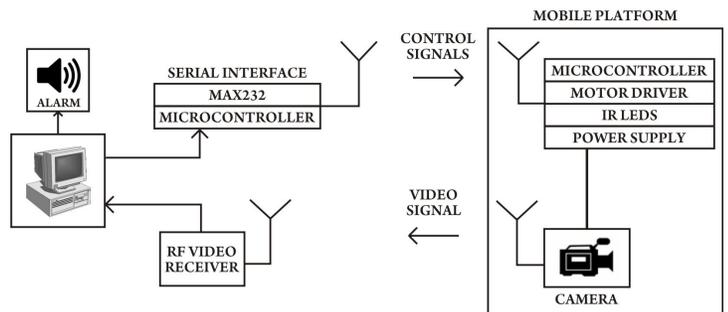

object to the control centre for analysis. A Program developed using MATLAB at the control centre is then

**Fig 4. Setup for motion detection mode**

used to analyze four consecutive and based on this analysis a security alarm is raised if required.

## 4. DEVELOPMENT OF CONTROLLING PROGRAMS
### 4.1 Program for mode 1

This program has been developed in Microsoft Visual Studio 6.0(Visual Basic programming language).It consists of 12 buttons, 2 checkboxes, 1 video box, 1 picture
box. These 16 buttons are configured as:
- 7 buttons to control the directions of the platform.
- 2 checkboxes for controlling lights and night vision respectively.
- 2 buttons for camera control (start, stop).
- 2 buttons for capturing video.
- 1 button for capturing snapshot

The video box displays the video using VideoCapPro ActiveX Control, received from the camera mounted on the mobile platform and similarly the picture box displays the snapshot taken when the button for capturing the snapshot is depressed. The program transmits the control signals via serial port using MSCOMM (Microsoft Common Control) component.

### 4.2 Program for mode 2

This program implements client-server architecture in VB using TCP and socket programming.

The client program (fig. 5) has total of 16 buttons,1 video box, 1 picture box and 2 text boxes which are configured as follows:
- 2 buttons for managing the connection between client and server.
- 1 text box to input the host IP address and other for communicating port.
- 7 buttons to control the directions of the mobile platform.
- 2 buttons for controlling lights and night vision respectively.
- 2 buttons for capturing video.
- 1 button for capturing snapshot.
- 2 buttons for camera control (start, stop).

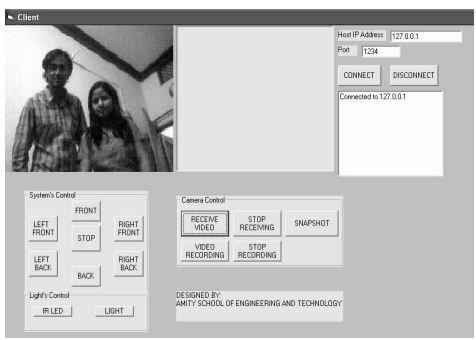

**Fig 5. A screen view of client program**

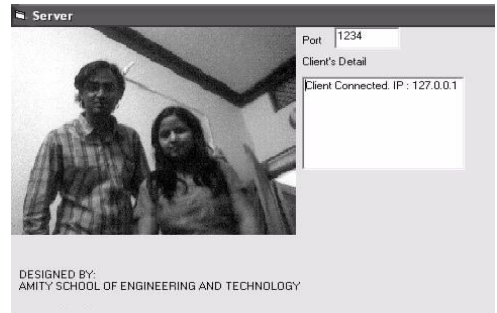

**Fig 6 A screen view of server program**

The server program (fig. 6) continuously listen to the port and establishes connection if requested. It then transmits the video to the client using VideocapX control and provides access to its serial port.

When connect button is depressed, client program connects to the server whose IP address has been specified in the IP text box at the given port using WINSOCK (Windows Socket) component. Once the connection is established, the server provides video feedback to the client. The client in turn controls the mobile platform via internet.

### 4.3 Program for mode 3

It has been developed in MATLAB (ver 7.01 from Math Works).The captured image is analyzed pixel by pixel. The screen is divided into four quadrants. Each pixel value is then compared with the stored color value. On comparing these values with the pixels of the captured image (fig. 7) of the object, those pixels are highlighted which matches with the specified color. The highlighted pixels (fig. 8) indicate the direction in which the mobile platform is to be moved. According to the quadrant of highlighted area, control signals are automatically transmitted to the mobile platform to track the object.

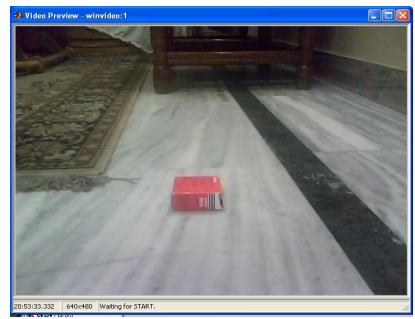

**Fig 7 unprocessed image**

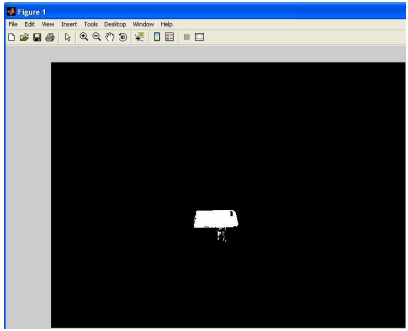

**Fig. 8 Processed image showing red color**

### 4.4 Program for mode 4

Again, MATLAB is used for programming. The GUI (fig. 9) has a button and one text box. The button incorporates multiple functions such as start camera, start and stop monitoring process. Monitoring can only be closed if correct password is entered in the text box. On activation the monitoring of scene in front of camera is performed continuously and visual information is sent to the control centre. At the control centre image processing is performed to detect any motion.

**4.5 Program for Mobile Platform** An AVR ATmega8 microcontroller, L298 motor driver IC is embedded in the mobile platform. The controller is programmed to drive the motor driver IC which in turn drives the motors of the mobile platform. AVR studio ver 4.0 is used as a program editor, WINAVR is used for compiling the code. The program is written into the microcontroller using Ponyprog (ver 1.17) which interfaces the programmer with the parallel port.

## 5. CONCLUSIONS

The designed advanced real time video motion detection system allows user to maneuver the mobile platform from a remote location. The system also provides feedback to the end user in terms of visuals rendering the system useful for spying purposes. This system integrates the functionality of four modes and selects one of them for unique application. The image processing discussed in mode 3 proves to be useful in industry applications and mode 4 can be useful for monitoring the highly restricted areas. Unlike most previous methods for real time video analysis the suggested approach used in mode 4 utilizes the difference between four consecutive frames, however, due to the unexpected traffic in the internet, delay in communication between the server and the end user may happen. This causes delays on executing commands and programs and transmitting on-site images.

## ACKNOWLEDGMENT

The first author Sumita Mishra is grateful to Prof. B. P. Singh, Maj. Gen. K.K. Ohri, Prof. S.T.H. Abidi and Brig. U. K. Chopra of Amity University, India for their support during the research work.